\documentclass{article}

\PassOptionsToPackage{numbers, compress}{natbib}
\usepackage[preprint]{neurips_2026}
\usepackage[utf8]{inputenc} % allow utf-8 input
\usepackage[T1]{fontenc}    % use 8-bit T1 fonts
\usepackage{hyperref}       % hyperlinks
\usepackage{url}            % simple URL typesetting
\usepackage{booktabs}
\usepackage{amsfonts}       % blackboard math symbols
\usepackage{nicefrac}       % compact symbols for 1/2, etc.
\usepackage{microtype}      % microtypography
\usepackage{xcolor}         % colors
\usepackage{amsmath}
\usepackage{xspace}

\usepackage{booktabs}       %tablemake
\usepackage{graphicx}
\usepackage{multirow}
\usepackage{multicol}
\usepackage{color}
\usepackage{colortbl}
\definecolor{Gray}{gray}{0.86}
\usepackage[most]{tcolorbox}
\usepackage{tabularx}

\usepackage{amssymb}
\usepackage{bbding}
\usepackage{pifont}
\usepackage{wasysym}
\usepackage{amssymb}

\usepackage{algorithm}
\usepackage{algorithmic}
\usepackage{amsmath, amssymb}
\usepackage{xcolor}
\newcommand{\algcomment}[1]{ \textcolor{blue!70!black}{\texttt{\#\ #1}}}

\usepackage{tikz}
\usetikzlibrary{tikzmark, calc, arrows.meta, fit, backgrounds}

\usepackage{subcaption}

\usepackage{svg}

\usepackage{pgfplots}

\usepackage{wrapfig}
\pgfplotsset{compat=1.18}

\title{dRAE: Representation Autoencoder with \\ Hyper-Spherical Codes}

\vspace{-1cm}
\author{%
    \begin{minipage}[t]{\textwidth}
    \centering
        \textbf{Tianren Ma}$^{1,4}$ \ \ \textbf{Lin Long}$^{2,4}$ \ \ \textbf{Chuyan Chen}$^{3,4}$ \ \ \textbf{Mu Zhang}$^1$ \\
        \vspace{0.1cm}
        \textbf{Junbo Zhao}$^{2,4*}$ \ \ \textbf{Tong Zhang}$^1$ \ \ \textbf{Qixiang Ye}$^{1}$\thanks{Corresponding authors} 
    \end{minipage}
}

\begin{document}

\maketitle

\vspace{-0.8cm}
\begin{center}
    $^1$ University of Chinese Academy of Sciences \ \
    $^2$ Zhejiang University \\
    $^3$ Peking University \ \
    $^4$ Ant Group \\
    \vspace{0.3cm}
    Project Page: \url{https://drae-hsq.github.io}
    % {\tt\small matianren18@mails.ucas.ac.cn  \ \ qxye@ucas.ac.cn}
\end{center}

\begin{abstract}
%

% Multimodal models require visual tokenizers that jointly capture high semantic density and fine-grained structural fidelity. Representation Autoencoders (RAEs) offer a promising direction by decoding images directly from a pretrained semantic space using high-dimensional continuous tokens, bypassing the information bottleneck of VAEs and benefiting both visual understanding and generation. 
In this work, we aim to discretize the high-dimensional visual representations to bridge the gap with language models — a non-trivial challenge, as existing quantization methods suffer from codebook collapse, failing to scale while preserving semantic coherence.
% However, discretizing these tokens for compatibility with language models faces a critical challenge: high-dimensional codebooks are prone to collapse, limiting scalability and degrading reconstruction quality. 
We identify the root cause as metric mismatch: standard Euclidean codebook objectives are fundamentally misaligned with the anisotropic geometry of representation space, leading to codebook embeddings with high-variance magnitude scales and uneven angular distributions that hinder scalability. To address this, we propose \textbf{Hyper-Spherical Quantization (HSQ)}, which decouples semantic content from feature magnitude via angular routing, preventing code assignment from being dominated by scale rather than meaning. The resulting discrete Representation Autoencoder (dRAE) achieves high-fidelity reconstruction while preserving semantic integrity and supporting scalable codebook budget. Extensive experiments demonstrate consistent performance gains as the vocabulary size scales to 131,072, along with 100\% codebook utilization, simplified training pipeline, and strong performance across understanding and generation tasks.
% Project page: \url{https://hsq-drae.github.io}

\end{abstract}

\section{Introduction}
The trajectories of visual understanding and generation have conventionally been separated by their representation designs. 
Visual encoders optimized for understanding~\cite{CLIPS2021,DINOv2,Siglip2025}, excel at mapping images to high-dimensional, semantically rich spaces via vision transformers (ViTs)~\cite{ViT2021,MAE2022}, whereas generative frameworks typically rely on variational autoencoders (VAEs)~\cite{VAE2014,SD-VAE2022,esserTamingTransformersHighResolution2021,DiT2023} to compress images into low-dimensional manifolds.
While modern VAEs are highly optimized for localized pixel reconstruction, scaling their latent dimensions is constrained by the KL-divergence prior and practical difficulties~\cite{vavae}.
Consequently, most \textit{unified} multimodal models resort to disjointed pipelines\textemdash employing a ViT for comprehension and an independent VAE for generation.

Recently, a promising trend has emerged to bridge this architectural gap. Frameworks~\cite{VILA-U2025,RAE2025} such as the representation autoencoder (RAE) demonstrate that high-dimensional features from contrastive or self-supervised models can be directly used to reconstruct images via a simple ViT.
% \TZ{To bridge this architectural divide, Representation Autoencoders (RAEs)~\cite{RAE2025} have emerged as a compelling alternative. By decoding images directly from high-dimensional, pretrained semantic spaces—leveraging features from contrastive or self-supervised models—RAEs produce visual tokens with significantly rich semantics than traditional VAEs. Because these representations inherently capture global semantic alignment, they naturally narrow the representation gap between visual and textual modalities.}
%
As these pretrained representations inherently capture global semantic alignment, visual tokens derived from this space carry significantly higher generalizable information than that from a pure reconstruction VAE. 
This high semantic bandwidth is desirable for multimodal models, as it minimizes the need for complex cross-modal alignment~\cite{scale-rae,Tong2026BeyondLM}, offering a more unified foundation for integrated understanding and generation.
% \TZ{Notably, the heavy optimization of VAEs for localized pixel reconstruction directly imposes a scalability bottleneck; by prioritizing low-level texture statistics under restrictive KL-divergence priors, these models produce ungeneralizable latents that fail to capture the semantic depth required for high-level understanding. This fragmentation forces modern multimodal models into disjoint pipelines, necessitating separate, independent architectures for comprehension and generation. For multimodal models, this high semantic bandwidth minimizes the need for complex cross-modal alignment, offering a more unified foundation for integrated understanding and generation.}

\begin{figure}
    \centering
    \includegraphics[width=1.0\linewidth]{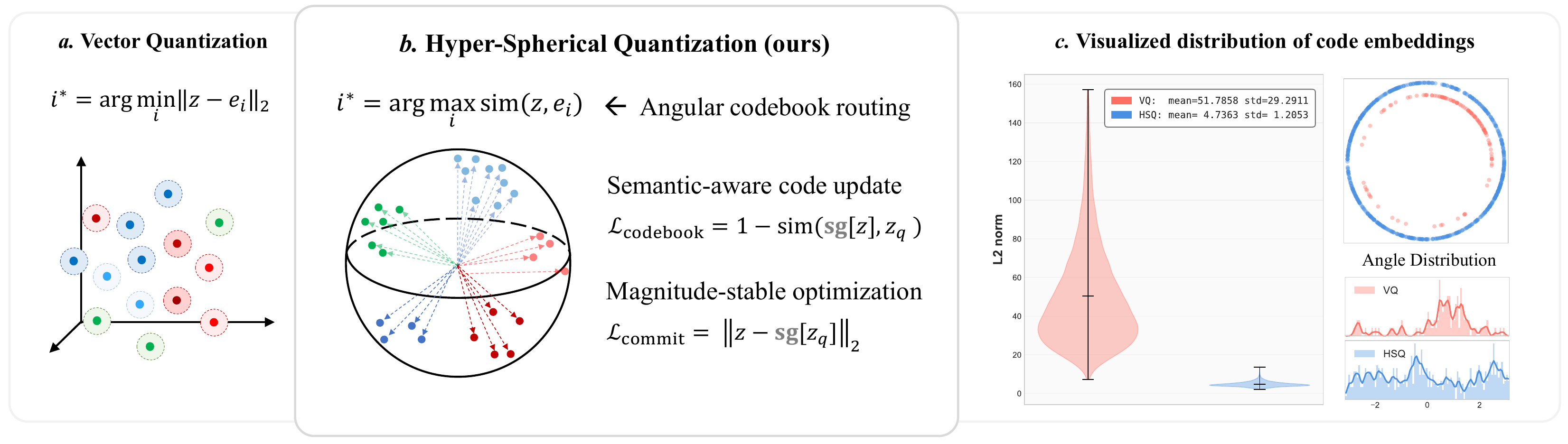}
    \caption{Illustration of the code space defined by VQ (\textit{a}) and HSQ (\textit{b}), and distribution of their learned codebook embeddings (\textit{c}). $\text{Sim}(\cdot)$ for cosine similarity, and $\text{sg}(\cdot)$ for stop-gradient operation. For the angular analysis in \textit{c}, we sample features from the central region with similar magnitude ranges and visualize them with PCA dimension reduction~\cite{isola-understandingclip}.}
    \label{fig:motivation}
\end{figure}

% HSQ decouples semantic content from feature magnitude via angular routing, preventing code assignment from being dominated by scale rather than meaning. It enables magnitude to encode low-lever details essential for image reconstruction.

However, transporting such signals via discrete indices presents a significant optimization challenge.
Existing vector quantization (VQ) methods~\cite{VQVAE2017}, Fig.~\ref{fig:motivation} \textit{a}, which are designed for low-dimensional autoencoder bottlenecks, typically result in severe codebook collapse when applied on high-dimensional features (see Fig.~\ref{fig:dynamics}).
Empirical observations in recent baselines~\cite{VQRAE2025} reveal that the tokenizer plateaus when scaled to a vocabulary larger than 16K codes.
This phenomenon neutralizes the primary advantage of representation-based designs, as the constrained vocabulary cannot adequately capture the diverse semantic capacity of the latent space.

In this study, we argue that the observed codebook collapse is largely attributable to the fundamental \textbf{metric mismatch}. 
While semantics in vision foundation models are predominantly encoded in the orientation of latent vectors~\cite{isola-understandingclip}, VQ-based frameworks rely on Euclidean distance for codebook assignment. This inconsistency makes the objective sensitive to the anisotropic scale of pretrained spaces, resulting in codebook embeddings with high-variance magnitude and clustered angular distributions, as evidenced in Fig.~\ref{fig:motivation} \textit{c}. 
Such magnitude-induced bias allows latent vectors with disparate norms to disproportionately dominate the tessellation, enabling certain codebook entries to "hijack" assignments based on numerical magnitude alone, regardless of their semantic alignment with the input.
At the same time, simply imposing a unit-sphere prior on inputs, which means discarding the magnitude, would deteriorate reconstruction, as this operation is non-uniform across tokens, leading to structural and textural nuances lost.

% \TZ{Deploying RAE-style representations within autoregressive frameworks necessitates discretization, yet this transition is where existing methods break down. Traditional vector quantization (VQ) methods suffer from severe codebook collapse when applied to these high-dimensional features, a failure we attribute to a fundamental metric mismatch. While semantics in vision foundation models are predominantly encoded in the orientation of latent vectors, standard VQ frameworks rely on Euclidean distance for codebook assignment. This inconsistency makes the objective hypersensitive to the anisotropic scale of pretrained spaces, resulting in codebook embeddings with high-variance magnitude scales and clustered angular distributions, as evidenced in Fig.~\ref{fig:motivation} Such magnitude-induced bias allows latent vectors with disparate norms to disproportionately dominate the tessellation, enabling certain codebook entries to "hijack" assignments based on numerical magnitude alone, regardless of their semantic alignment with the input. This bottleneck effectively neutralizes the advantage of representation-based designs, causing the tokenizer to plateau when scaled beyond 16K codes. However, simply imposing a unit-sphere prior to eliminate this bias is insufficient; discarding feature magnitude altogether deteriorates reconstruction quality, as critical low-level structural and textural nuances are lost in the process.}.

To resolve this dilemma, we propose the discrete Representation Autoencoder (dRAE) powered by Hyper-Spherical Quantization (HSQ). The core innovation lies in explicitly decoupling the metric used for codebook routing from the quantization objective. By shifting from Euclidean distance to angular similarity, HSQ assigns codes based on semantic orientation while preserving essential magnitude information for the decoder. This approach significantly improves tokenization quality through two primary mechanisms: \textbf{magnitude homogenization} and \textbf{angular uniformity}. As illustrated in Fig.~\ref{fig:motivation} \textit{c}, HSQ constrains the codebook to a significantly more stable magnitude scale, effectively eliminating the high-variance outliers that trigger codebook collapse in vanilla VQ or SimVQ~\cite{simvq}. By reducing the magnitude-induced noise, the latent space attains a more isotropic, uniform distribution on the hyper-sphere. This ensures that the codebook utilizes its full capacity to represent distinct semantic directions rather than being dominated by arbitrary feature scales. Consequently, dRAE preserves the fine structural details necessary for high-fidelity synthesis while maintaining the semantic integrity required for large-scale multimodal understanding.

Through extensive experiments, we demonstrate that the proposed HSQ not only shows continuous gain on reconstruction by scaling vocabulary up to $131,072$, but also keeps high semantic fidelity for image understanding tasks.
We also validate the practical efficacy of dRAE by integrating it with a generative pipeline for text-to-image generation, confirming its viability as a high-throughput tokenizer for both visual understanding and generation tasks.

% In a nutshell, our contributions can be summarized:
% \begin{itemize}
%     \item We identify metric mismatch — the misalignment between Euclidean codebook objectives and the anisotropic geometry of pretrained representation spaces — as the root cause of codebook collapse in high-dimensional discrete tokenization.
%     \item We propose Hyper-Spherical Quantization (HSQ), an angular routing scheme that decouples codebook assignment from feature magnitude, enabling stable and scalable quantization of high-dimensional visual features.
%     \item We present dRAE, a discrete Representation Autoencoder that scales cleanly to a vocabulary of 131,072 codes with consistent performance gains across image reconstruction, visual understanding, and text-to-image generation.
% \end{itemize}

\section{Related Work}

\noindent\textbf{Visual Semantic and Generative Latent.} 
The evolution of self-supervised and contrastive learning has established a robust foundation for multimodal models. By utilizing objectives such as masked modeling~\cite{MAE2022}, self-distillation~\cite{DINO,DINOv2} and image-text alignment~\cite{CLIPS2021,Siglip2025}, visual models embed image patches into high-dimensional latents (\textit{e.g.} $d \in [768, 1536]$), and learn semantically structured  representation that generalizes across visual understanding tasks.
%
% While these spaces excel at high-level understanding, they are not inherently designed for detail-persevering reconstruction.
%
On the other hand, state-of-the-art generative models still operate on heavily compressed manifolds ($d \in [8, 64]$), \textit{i.e.}, diffusion is typically built in reconstruction-trained VAE space~\cite{SD-VAE2022, vavae}.
This creates a dimensional incompatibility between the semantic-rich embeddings and the compact generative latents.
%
% Recent studies attempts to augment latent quality through representation-VAE alignment training~\cite{REPA-E2025, vavae}, but these methods introduce extra training stages and/or objective complexity.
%

\noindent\textbf{Unifying Comprehension and Generation.} 
One direct solution of aligning these disparate paths is to explore an ensemble of semantic and reconstruction objectives. VTP~\cite{VTP2025} explores end-to-end pretraining to learn representations that can unify comprehension with generation from scratch. However, it imposes extremely high demand on data collection and training budgets. A more promising trend is to utilize a pretrained encoder for more downstream tasks. VILA-U~\cite{VILA-U2025} uses residual quantization to align image latents with text embeddings, while TokLIP~\cite{TokLIP2025} bakes CLIP features into the pretrained VQGAN tokenizer through distillation and contrastive training.
Recently, RAE~\cite{RAE2025} proposes that the high-dimensional features, optimized for semantics, can be directly decoded for image reconstruction. It also shows that high-dimensional vectors, which jointly capture high semantic density and fine-rained structural, are of great potential in visual world modeling~\cite{Tong2026BeyondLM}.
Its discrete counterpart, VQRAE~\cite{VQRAE2025}, uses SimVQ~\cite{simvq} to quantize the high-dimensional tokens, but is prone to codebook collapse and of limited scaling potential.
% Recent efforts aim to align these two disparate paths. VILA-U~\cite{VILA-U2025} uses residual quantization to align image latents with text embeddings. 
% %
% TokLIP~\cite{TokLIP2025} bakes CLIP features into the pretrained tokenizer through distillation and contrastive training.
% %
% VTP~\cite{VTP2025} explores end-to-end pretraining to learn representations that can unify   comprehension with generation tasks.
% %
% RAE~\cite{RAE2025} attempts to learn high-dimensional token vectors that jointly capture high semantic density and fine-rained structural, by using encoders pretrained for classification to reconstruct images.
% %
% Its discrete counterpart, VQRAE~\cite{VQRAE2025}, introduces an additional distillation loss for finetuning, yet its codebook size is still constrained, showing limited benefits when scaled up.
%
In this study, we attempt to address these limitations by introducing a scalable quantization method that stabilizes training while preserving semantic depth.

\section{Preliminary}

\paragraph{Image Reconstruction.}
This requires the model to learn a mapping from the input image $x \in [0,1]^{h \times w \times 3}$ to latents and reconstruct them back to pixel space. Formally, an encoder $E$ maps $x$ to latent features $Z := E(x) = \{z_n\}_{n=1}^N \in \mathbb{R}^{N \times d}$, where $N = hw/p^2$ denotes the number of visual tokens. We will also use $z\in\mathbb{R}^{d}$ to denote an un-quantized embedding. A decoder $D$ reconstructs the image $\hat{x} = D(Z)$. The model is trained to minimize a objective that encourages both pixel-level fidelity ($\text{L}_1$ distance) and perceptual quality (perceptual $\mathcal{L}_{\text{perc}}$ and discriminator loss $\mathcal{L}_{\text{disc}}$):
\begin{equation}
    \mathcal{L}_{\text{rec}} = \text{L}_1(\hat{x}, x) + \omega_{\text{p}}  \mathcal{L}_{\text{perc}}(\hat{x}, x) + \omega_{\text{d}} \mathcal{L}_{\text{disc}}(\hat{x}, x).
\end{equation}

\paragraph{Representation Autoencoder.}
Conventional approaches such as VAEs compress images into low-dimensional latent spaces, $i.e.$, the bottleneck, which facilitates convergence but limits the capacity to encode semantic information. In contrast, RAE removes the bottleneck design, and instead leverages high-dimensional, semantically structured embeddings directly from pretrained Vision Foundation Models (VFMs) as its latents. 
Specifically, RAE employs a frozen VFM as $E$ and a trainable ViT as $D$. By avoiding aggressive compression, RAE maintains features that are beneficial for downstream visual understanding and generation tasks. 

\paragraph{Vector-Quantizing Semantics.}
VFM features can be quantized for discrete modeling. Let $Q(\cdot)$ denote the quantization operation, the quantized features as $z_q:= Q(z)$ and $Z_q := Q(Z)$. VQRAE projected the image features to 1536 dimensions, and each vector $z_n$ is mapped to its nearest entry from a learnable codebook $\mathcal{C} = \{c_i\}_{i=1}^K$, where $K$ is the codebook capacity:
\begin{equation}
    z_q = \arg\min_{c_i \in \mathcal{C}} \| z - c_i \|_2.
\end{equation}
The quantization objective consists of a codebook loss and a commitment loss:
\begin{equation}
\label{eq:codebook_loss}
    \mathcal{L}_{\text{VQ}} = \underbrace{\| \text{sg}[z] - z_q \|_2^2}_{\mathcal{L}_{\text{codebook}}} + \beta \underbrace{\| z - \text{sg}[z_q] \|_2^2}_{\mathcal{L}_{\text{commit}}},
\end{equation}
where $\text{sg}[\cdot]$ denotes the stop-gradient operator and $\beta$ controls the commitment strength.

To preserve the semantic structure while improving reconstruction fidelity, VQRAE also adopts a self-distillation strategy. A teacher model $T$ (initialized from the original VFM) provides supervision to the unfrozen $D$, encouraging the learned features $Z_I$ to remain close to the teacher representations:
\begin{equation}
    \mathcal{L}_{\text{total}} = \mathcal{L}_{\text{rec}} + \mathcal{L}_{\text{VQ}} + \lambda \| Z - T(x) \|_2^2.
\end{equation}

Additionally, to improve codebook utilization, a learnable projection matrix $W \in \mathbb{R}^{d \times d}$ is introduced by SimVQ~\cite{simvq} to transform codebook entries as $Wc_i$, allowing for a more flexible alignment.

\section{Methodology}

\subsection{Anisotropy in High-dimensional Space}

In high-dimensional Euclidean space, a standard Gaussian distribution exhibits the \textit{thin-shell} concentration property~\cite{vershynin_high-dimensional_2018, Ledoux2001TheCO}: if $z \sim \mathcal{N}(0, I_d)$, its norm $\|z\|_2$ concentrates sharply around $\sqrt{d}$ as $d \to \infty$. 
Vision encoders such as CLIP~\cite{CLIPS2021} are trained with contrastive objectives that encourage features to spread uniformly across directions-a pressure analogous to the isotropic structure of a Gaussian~\cite{betserInfoNCEInducesGaussian2026}. Related self-supervised approaches like DINO~\cite{DINO,DINOv2} series and JEPA~\cite{assranSelfSupervisedLearningImages2023} also yield highly regular representations~\cite{balestrieroGaussianEmbeddingsHow2025}, which have been shown to encode density-related structure that can be exploited with Gaussian models. Their patch embeddings, as suggested by recent studies~\cite{ellipsoidclip, isola-understandingclip}, similarly concentrate on a thin spherical shell, with meaningful information encoded in both the direction (angular) and the radius (magnitude) of each feature vector. We conduct diagnostic experiments to empirically investigate their function, with the aim of validating two hypotheses:

\textbf{H1.} \textit{Semantic information roughly lies on a hyper-sphere for understanding.} We train two variants of an MLLM following LLaVA's ViT-MLP-LLM architecture~\cite{llava-1.5}. The original version receives raw image features $Z$ as MLP's inputs. In the normalized version, we apply $\ell_2$ normalization such that all visual features $Z' = Z/\|Z\|_2$ reside on a unit hyper-sphere. 
As shown in Tab.~\ref{tab:mllm_results}, the performance gap across diverse benchmarks—ranging from visual perception (MMBench~\cite{mmbench}) to hallucination (POPE~\cite{pope}) —is negligible ($\sim1\%$). This confirms that the semantic signal is almost entirely preserved in the feature's directional orientation for high-level understanding.

\textbf{H2.} \textit{Magnitude is structurally essential for image reconstruction.} Following RAE, we train a ViT decoder to reconstruct pixels from either $Z$ or $Z'$. As shown in Tab.~\ref{tab:recon_results}, models trained on normalized features struggle to recover global structure, resulting in worse performance on reconstruction metrics.

\begin{table}[t]
    \centering
    \caption{\textbf{Diagnostic Analysis.} Comparison of understanding (a) and reconstruction (b) performance between raw and normalized image features as inputs.}
    \label{tab:diagnostic_main}
    \vspace{2mm}
    \begin{minipage}{0.50\textwidth}
        \centering
        \subcaption{Multimodal Understanding}
        \label{tab:mllm_results}
        \small
        \begin{tabular}{l|ccc}
            \toprule
            Variant & MMBench $\uparrow$ & TextVQA $\uparrow$ & POPE $\uparrow$\\
            \midrule
            Raw $Z$ & 82.2 & 61.3 & 85.2\\
            Norm $Z'$ & 81.1 & 60.7 & 84.3\\
            \bottomrule
        \end{tabular}
    \end{minipage}
    \hfill
    \begin{minipage}{0.45\textwidth}
        \centering
        \subcaption{Image Reconstruction Fidelity}
        \label{tab:recon_results}
        \small
        \begin{tabular}{l|cc|c}
            \toprule
            Variant & PSNR $\uparrow$ & SSIM $\uparrow$ & rFID $\downarrow$ \\
            \midrule
            Raw $Z$ & 22.5 & 0.62 & 4.62 \\
            Norm $Z'$ & 20.6 & 0.51 & 9.57 \\
            \bottomrule
        \end{tabular}
    \end{minipage}
\end{table}

The diagnostic results suggest that, to build a tokenizer that captures both high semantic density and fine-grained structural fidelity, we need a scalable quantization algorithm that preserves both the angular and magnitude information of high-dimensional features.

\subsection{Hyper-Spherical Quantization}\label{sec:HSQ}

\begin{minipage}[b]{0.46\linewidth}

\paragraph{The Unreliable Euclidean. }As discussed above, representation features approximately concentrate on a hyper-spherical shell rather than a full Euclidean space. However, VQ optimizes codebooks using the $\ell_2$ distance, as in Eq.~\ref{eq:codebook_loss}. Such an objective is linearly sensitive to variations in feature norms, which biases the partitioning toward radial differences rather than angular structure. Consequently, samples with larger or varying norms can disproportionately influence the tessellation, irrespective of their directional alignment, as shown in Fig.~\ref{fig:motivation}. We posit that this mismatch is a primary cause of the codebook collapse phenomenon observed in Fig.~\ref{fig:dynamics}.
% At the same time, simply discarding the magnitude and operate on unit sphere would impair reconstruction performance, evidenced in Tab.~\ref{tab:recon_results}.

\end{minipage}%
\hfill
\begin{minipage}[b]{0.52\linewidth}
    \centering
    \includegraphics[width=\linewidth]{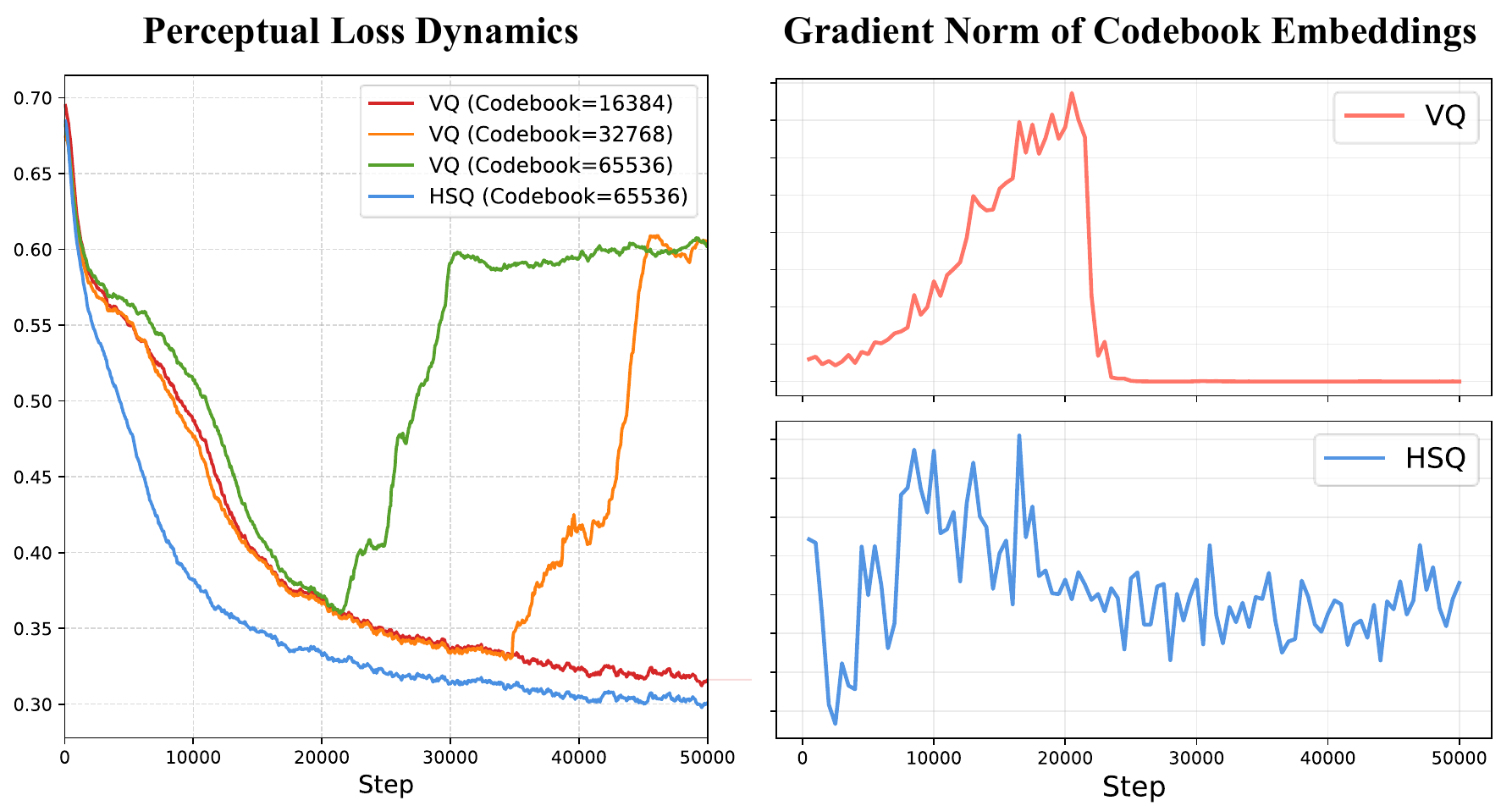}
    \captionof{figure}{Left: Training curve under identical settings. Right: Gradient norm of codebook during training, where VQ's collapse mode can be observed.}
    \label{fig:dynamics}
\end{minipage}

\begin{figure}[h]
\begin{tikzpicture}[remember picture, overlay]
    
    % --- Left Side: Original Algorithm ---
    \node (original) at (0,0) [anchor=north west] {
        \begin{minipage}{0.5\textwidth}
        \begin{algorithm}[H]
        \caption{Classic Vector Quantization}
        \begin{algorithmic}[1]
        \small
        \STATE{\textbf{Input: } $Z \in \mathbb{R}^{N\times d}, C, \beta$}\vspace{0.5em}
        
        % Mark the lines for the highlight box
        \STATE \tikzmarknode{start_zoom}{$D_{i,j} \gets \|Z_i - C_j\|^2_2$}\vspace{0.4em}
        \STATE $I_i \gets \arg\min_j D_{i,j}$ \vspace{0.4em}
        \STATE $Z_q \gets C[I]$\vspace{0.4em}
        \STATE \tikzmarknode{end_zoom}{$\mathcal{L}_{\text{codebook}} \gets \|\text{sg}[Z] - Z_q\|^2_2$}\vspace{0.8em}
        
        \STATE $\mathcal{L}_{\text{commit}} \gets \|Z - \text{sg}[Z_q]\|^2_2$\vspace{0.2em}
        \STATE $\mathcal{L}_{\text{VQ}} \gets \mathcal{L}_{\text{codebook}} + \beta \mathcal{L}_{\text{commit}}$\vspace{0.2em}
        % \STATE $Z_{\text{out}} \gets Z + \text{sg}[Z_q - Z]$\vspace{0.5em}
        % \STATE{\textbf{Return: }$Z_{\text{out}}, \mathcal{L}_{\text{VQ}}, I$}
        \STATE{\textbf{Return: }$\mathcal{L}_{\text{VQ}}$}
        \end{algorithmic}
        \end{algorithm}
        \end{minipage}
    };

    % --- Right Side: Zoomed Section (Angular Logic) ---
    \node (zoomed) at (7.5, -0.4) [
        anchor=north west, 
        draw=blue!40, 
        line width=1pt, 
        rounded corners=2pt, 
        fill=blue!5,
        inner xsep=12pt,
        inner ysep=10pt
    ] {
        \begin{minipage}{0.38\textwidth}
            {\textbf{\small \color{blue!70!black} Hyper-Spherical Quantization} \par} 
            \vspace{0.8em}
            \small
            \algcomment{Compute cosine similarity}\\
            \textbf{3:} $S_{i,j} \gets \frac{Z_i}{\|Z_i\|_2 } \cdot \frac{C_j}{ \|C_j\|_2}$ \\[0.1em]
            
            \algcomment{Angular routing}\\
            \textbf{4:} $I_i \gets \arg\max_j S_{i,j}$ \\
            \textbf{5:} $Z_q \gets {C}[I]$ \\[0.1em]
            
            \algcomment{Spherical codebook loss}\\
            \textbf{6:} $\mathcal{L}_{\text{codebook}} \gets 1-\text{sg}[\frac{Z}{\|Z\|_2}] \cdot \frac{Z_q}{ \|Z_q\|_2}$
        \end{minipage}
    };

    % --- The "GitHub Style" Highlight & Connectors ---
    \begin{scope}[on background layer]
        % 1. The Light Gray Highlight Box on the left
        \node[
            fill=gray!12, 
            draw=gray!60, 
            densely dotted, 
            line width=0.8pt,
            inner sep=4pt,
            rounded corners=1pt,
            fit=(start_zoom) (end_zoom)
        ] (highlight) {};

        % 2. Subtle connection shading (the "zoom" fan)
        \fill[blue!10, opacity=0.4] 
            (highlight.north east) -- (zoomed.north west) -- 
            (zoomed.south west) -- (highlight.south east) -- cycle;
            
        % 3. Clean boundary lines for the fan
        \draw[blue!20, line width=0.5pt] (highlight.north east) -- (zoomed.north west);
        \draw[blue!20, line width=0.5pt] (highlight.south east) -- (zoomed.south west);
    \end{scope}

\end{tikzpicture}
\label{comparison}
\vspace{5cm}
\end{figure}

\paragraph{Objective Decoupling. } To mitigate these challenges, we propose Hyper-Spherical Quantization (HSQ). Our core insight is to decouple the semantic orientation of a feature from its magnitude during the quantization step. Following VQRAE, we add a learnable linear projection on the raw image features, then, HSQ replaces the standard Euclidean assignment with angular routing:
\begin{equation}
    I_i = \arg\max_j \frac{Z_i \cdot {C}_j}{\|Z_i\|_2 \| {C}_j\|_2}.
\end{equation}

By transitioning to an angular-based metric, we assign the codebook based on purely the directional relationships learned during pretraining. To maintain this constraint, we also rewrite the codebook loss $\mathcal{L}_{\text{codebook}}$ as a spherical objective, which constrains codebook updates to the tangent space. 
\begin{equation}
    \mathcal{L}_{\text{codebook}} = 1 - \text{sg}\left[\frac{Z}{\|Z\|_2}\right] \cdot \frac{Z_q}{\|Z_q\|_2}.
\end{equation}

% as illustrated by the red curve in Fig.~\ref{fig:analysis}. This distribution exhibits a more expansive dynamic range, effectively "spreading" the features across a manifold of constant curvature.
% As a result, HSQ effectively "resurrectes" unused codes, enabling stable scaling of the vocabulary size to 130K entries without suffering from collapse.

\paragraph{How is the magnitude optimized? }
The magnitude scaling of code vectors is not arbitrary. In fact, replacing all objectives with angular-based methods leads to unstable and slow convergence, since the magnitude is completely unsupervised (see Sec.~\ref{sec:ablation} for details). Therefore, we retain the Euclidean objective for commitment loss deliberately, as it provides magnitude-aware guidance to the projected features. The core comparison is shown above. Please refer to Append.~\ref{appen:details} for the complete training algorithm, where full implementation like the straight-through estimator (STE) and codebook projection are included.

\subsection{Distinction from Existing Quantization Schemes}
\label{sec:distinctions}

To fully appreciate our design, it is instructive to compare HSQ with two popularized quantization paradigms: Finite Scalar Quantization (FSQ)~\cite{FSQ2024} and Spherical Vector Quantization (SVQ)~\cite{SVQ2025}.
 
FSQ dispenses with a learnable codebook entirely, instead bounding the continuous latents via a squashing function (\textit{e.g.}, $\tanh$) and rounding them to a predefined discrete grid. While FSQ effectively eliminates collapse by removing competitive codebook learning, it is confined to low dimensions. The implicit vocabulary size in FSQ scales exponentially with the dimension $d$ as $K = L^d$, where $L$ is the number of bins per dimension. This will yield \textit{intractable} vocabulary size for high-dimensional features. In contrast, HSQ operates directly in the native high-dimensional space while maintaining a highly utilized codebook.

SVQ projects latents onto a unit sphere, but it traditionally do so to impose a uniform yet strict prior on the latent space for training generative models from scratch. Consequently,it passes the \textit{normalized} vectors to the decoder. Crucially, HSQ differs by explicitly decoupling the angular routing from the magnitude-aware reconstruction, which does not enforce SVQ's unit sphere prior on the inputs. As detailed in Sec.~\ref{sec:HSQ}, we only use the spherical geometry during the nearest-neighbor assignment. For the actual representation passed to the decoder (and used in the commitment loss), we retrieve the \textit{unnormalized} code vector. 

% \begin{algorithm}[H]
% \caption{Hyper-Spherical Quantization}
% \label{alg:spherical_vq}
% \begin{algorithmic}[1]

% \STATE{\textbf{Input: }Feature tensor $Z \in \mathbb{R}^{N\times d}$. 
% Base codebook embeddings $\mathcal{C}$, learnable projection $W$, and commitment cost $\beta$} \\
% \vspace{0.3em}
% \STATE $\hat{C} \gets W\mathcal{C}$
% \algcomment{Project codebook following SimVQ}\\
% \STATE $S_{i,j} \gets \frac{Z_i}{\|Z_i\| } \cdot \frac{\hat C_j}{ \|\hat C_j\|}$
% \algcomment{Compute cosine similarity}\\
% \STATE $I_i \gets \arg\max_j S_{i,j}$
% \algcomment{Angular routing}\\
% \STATE $Z_q \gets \hat{C}[I]$
% \algcomment{Lookup on projected codebook}\\
% \STATE $\mathcal{L}_{\text{codebook}} \gets 1-\text{sg}[\frac{Z}{\|Z\| }] \cdot \frac{Z_q}{ \|Z_q\|}$
% \algcomment{Angular codebook loss}\\
% \STATE $\mathcal{L}_{\text{commit}} \gets \|Z - \text{sg}[Z_q]\|^2_2$
% \algcomment{Commitment loss}\\
% \STATE $\mathcal{L}_{vq} \gets \mathcal{L}_{\text{codebook}} + \beta \mathcal{L}_{\text{commit}}$
% \STATE $Z_{\text{out}} \gets Z + \text{sg}[Z_q - Z]$
% \algcomment{Straight-through estimator}\\
% \vspace{0.5em}

% \STATE{\textbf{Return: }$Z_{\text{out}}, \mathcal{L}_{vq}, I$}
% \end{algorithmic}
% \end{algorithm}

\section{Experiment}
% \textit{(Note: This section will be expanded with the final large-scale experimental results.)}

\subsection{Setups}
We provide a brief overview of the implementation in this section. Full experimental details and hyperparameters are included in Sec.~\ref{appen:details}. Throughout the experimental section, VQ denotes vector quantization trained with the conventional $\ell_2$ objective. Unless specified, all VQ implementations follow the SimVQ design, where the codes are projected with an linear projection layer.

\noindent\textbf{Implementation.} Following VQRAE, we utilize a pretrained SigLIP2 ViT-So400M~\cite{Siglip2025} as the image encoder, and use a ViT of symmetric design for the decoder. Unlike previous methods~\cite{VQRAE2025, TokLIP2025} that may use two-stage training for stability and performance trade-off, dRAE is trained end-to-end with distillation loss in one stage. With no specific embedding initialization, anti-collapse trick nor curriculum learning, the training pipeline is much simplified. 
% We unfreezing the encoder to refine pixel-level details while preserving high-level semantics with teacher distillation. 

For image understanding, we employ Qwen2.5-7B~\cite{baiQwen25VLTechnicalReport2025} as the LLM backbone, and adapt the commonly used visual instruction tuning pipeline as in LLaVA-1.5~\cite{llava-1.5}. For text-to-image (T2I) generation, we use Flan-T5-XXL~\cite{flan-t5} for text conditioning, and a diffusion transformer with discrete prediction head~\cite{maReDDiTRehashingNoise2025a, sahooSimpleEffectiveMasked2024}. The generative training follows the discrete diffusion paradigm, where images are tokenized into discrete indices and masked randomly during training. The training process is optimized via a time-weighted masked cross-entropy loss. We also train several class-to-image (C2I) generative variants on ImageNet for ablative study. See ~\ref{appn:mllm} for details. 

\noindent\textbf{Evaluation Metrics.} We evaluate reconstruction quality using reconstruction FID (rFID) ~\cite{gFIDheusel2017gans}, Peak Signal-to-Noise Ratio (PSNR), and Structural Similarity Index Measure (SSIM) on the ImageNet-1K~\cite{imagenet} validation set. For multimodal understanding, we evaluate on GQA~\cite{gqa}, TextVQA (TQA)~\cite{textvqa}, MMBench-en (MMB)~\cite{mmbench}, MME-Perception (MME-P)~\cite{mme}, and SEEDBench-Img (SEED)~\cite{seedb}. For C2I generation, we evaluate using generation FID (gFID) and Inception Score (IS)~\cite{IS2016}. For T2I generation, we evaluate on GenEval~\cite{geneval} and DPG-Bench~\cite{dpgbench}.

\subsection{Image Reconstruction}\label{sec:recon}

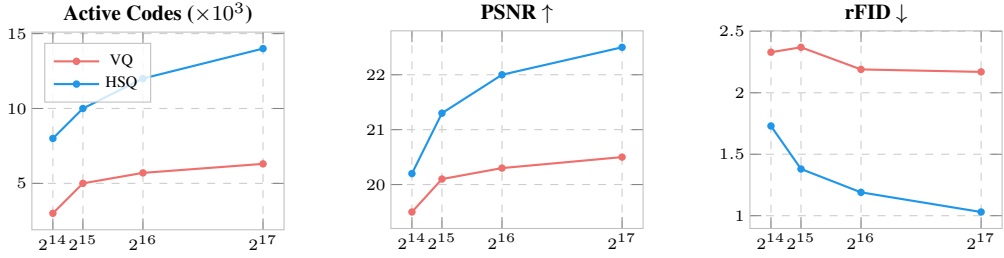
\begin{figure}[t!]
\centering

% Soft Modern Colors (Bright but professional)
\definecolor{softBlue}{RGB}{30, 150, 233}  % Clear Sky Blue
\definecolor{softCoral}{RGB}{239, 111, 108} % Soft Coral/Salmon

\pgfplotsset{
    neurips_style/.style={
        width=1.1\linewidth, 
        height=4.2cm,
        scaled x ticks=false,
        xtick={16384, 32768, 65536, 131072},
        xticklabels={$2^{14}$, $2^{15}$, $2^{16}$, $2^{17}$},
        grid=major,
        grid style={dashed, gray!40},
        axis line style={gray!50},
        tick label style={font=\tiny},
        title style={font=\footnotesize\bfseries, yshift=-5pt},
        label style={font=\tiny},
        % --- THE LEGEND FIX ---
        legend style={
            font=\tiny, 
            draw=gray!50,       % Thin border
            fill=white,         % Background color
            fill opacity=0.8,   % Slightly transparent
            text opacity=1,     % Ensure text is fully visible
            row sep=-1pt        % Compact spacing
        },
        % ----------------------
        enlarge x limits=0.1,
        inner sep=0pt,
        outer sep=2pt,
    }
}

% Panel 1
\begin{minipage}{0.32\linewidth}
\centering
\begin{tikzpicture}[trim axis left, trim axis right]
\begin{axis}[
    neurips_style,
    title={Active Codes ($\times 10^3$)},
    legend pos=north west,
]
\addplot[softCoral, mark=*, mark size=1pt, thick] coordinates {(16384,3) (32768,5) (65536,5.7) (131072,6.3)};
\addlegendentry{VQ}
\addplot[softBlue, mark=*, mark size=1pt, thick] coordinates {(16384,8) (32768,10) (65536,12) (131072,14)};
\addlegendentry{HSQ}
\end{axis}
\end{tikzpicture}
\end{minipage}
\hfill
% Panel 2
\begin{minipage}{0.32\linewidth}
\centering
\begin{tikzpicture}[trim axis left, trim axis right]
\begin{axis}[
    neurips_style,
    title={PSNR $\uparrow$},
]
\addplot[softCoral, mark=*, mark size=1pt, thick] coordinates {(16384,19.5) (32768,20.1) (65536,20.3) (131072,20.5)};
\addplot[softBlue, mark=*, mark size=1pt, thick] coordinates {(16384,20.2) (32768,21.3) (65536,22.0) (131072,22.5)};
\end{axis}
\end{tikzpicture}
\end{minipage}
\hfill
% Panel 3
\begin{minipage}{0.32\linewidth}
\centering
\begin{tikzpicture}[trim axis left, trim axis right]
\begin{axis}[
    neurips_style,
    title={rFID $\downarrow$},
]
\addplot[softCoral, mark=*, mark size=1pt, thick] coordinates {(16384,2.33) (32768,2.37) (65536,2.19) (131072,2.17)};
\addplot[softBlue, mark=*, mark size=1pt, thick] coordinates {(16384,1.73) (32768,1.38) (65536,1.19) (131072,1.03)};
\end{axis}
\end{tikzpicture}
\end{minipage}

\caption{\textbf{Scaling behavior with $K=2^{14} \sim 2^{17}$ vocabulary size.} From left to right: active code per batch, reconstruction PSNR, and rFID. }
\label{fig:vocab_scaling}
\end{figure}

Unlike prior observations that suggest limited gains from increasing vocabulary size~\cite{VQRAE2025}, the proposed approach demonstrates consistent scaling behavior. As shown in Fig.~\ref{fig:vocab_scaling}, increasing the vocabulary size leads to higher code utilization, which correlates with improved reconstruction fidelity.
As a contrast, the VQRAE (VQ) baseline demonstrates marginal improvement of active code number and PSNR when scaling up the vocabulary size.
Note that the active code number is calculated by summing up the unique code usage per batch during training.
%
%This experiments show the scaling behavior of vocab

Our hyper-spherical code design exhibits strong resistance to codebook collapse. Under a vocabulary size of $K=65{,}536$ with batch size 64 per GPU, the VQRAE baseline only activates $\sim$6K codes per forward pass, whereas our method consistently more than $\sim$12K codes, while keeping a high global code utilization ratio ($>90\%$) without stochastic sampling tricks. For HSQ, larger vocabularies also yield consistently better reconstruction performance, while VQ method's improvement (before collapse) is relatively marginal.
As can be seen in Tab~\ref{tab:rec}, under the same codebook budget ($16,384$), dRAE outperforms VQRAE with significant margins. With a $131,072$ codebook, dRAE achieves 0.42 rFID, and its PSNR and SSIM are also the best among the compared discrete tokenizers.

\begin{table*}[ht]
\centering
\caption{\textbf{Evaluation on reconstruction quality. }NA refers to direct utilization of pretrained features. Dim. indicates latent dimension (bottleneck) for autoencoders.  *: See~\ref{appen:eval} for evaluation details.}
\label{tab:rec}
\resizebox{0.83\textwidth}{!}{%
\begin{tabular}{lcccccc}
\toprule
Autoencoders & Method & Codebook & Dim. & rFID $\downarrow$ & PSNR $\uparrow$& SSIM $\uparrow$ \\
\midrule
\multicolumn{6}{l}{\small\textit{Low-dimensional}} \\
SD-VAE~\cite{SD-VAE2022} & VAE & -- & 32 & 0.49 & 26.10 & 0.72\\
UniLIP*~\cite{unilip} & VAE & -- & 32 & 0.79 & 22.99 & 0.67 \\
MUSE-VL~\cite{muse-vl} & VQ & $32768$ & 8 & 2.26 & 20.14 & 0.65 \\
QLIP~\cite{Qlip} & BSQ & $2^{28}$ & 28 & 3.21 & 23.16 & 0.63 \\
\midrule
\multicolumn{6}{l}{\small\textit{Representational}} \\
RAE~\cite{RAE2025} & NA & -- & 768 & 0.62 & 19.20 & 0.44\\
VILA-U~\cite{VILA-U2025} & RQ & $16384$ & 1152 & 1.25 & -- & -- \\
VQRAE*~\cite{VQRAE2025} & VQ & $16384$ & 1152 & 1.31 & 22.33 & 0.67 \\
\textbf{dRAE} & HSQ & $16384$ &1152 & \bf 0.69 & \bf 24.03 & \bf 0.70 \\
\textbf{dRAE} & HSQ & $131072$ &1152 & \bf 0.42 & \bf 24.52 & \bf 0.72 \\
\bottomrule
\end{tabular}
}
\label{tab:recon_comparison}
\end{table*}

\subsection{Multimodal Tasks}
\noindent\textbf{Image Understanding. }
dRAE demonstrates competitive performance on multimodal understanding benchmarks. We follow VQRAE to evaluate the tokenizer's performance using its tuned image encoder. Note that VQRAE is trained with original encoder but tested with the tuned one, while our method uses the encoder optimized for reconstruction throughout MLLLM training and testing, consistent with other baselines. As shown in Tab.~\ref{tab:und}, while the tokenizer training is end-to-end, dRAE maximally preserved the semantic knowledge encoded in the pretrained encoder. 
% In contrast, even when applying the same distillation strategy, VQ still lag behind HSQ-derived tokenizers, as shown in Tab.~\ref{tab:ablation_und}.

\begin{table*}[t]
\centering
\small
\caption{\textbf{Evaluation on multimodal understanding}. Res. for image resolution. All the models listed use $\sim7$B LLM as the backbone. Top-two best results are highlighted.}
\label{tab:und}
\resizebox{0.85\textwidth}{!}{%
\begin{tabular}{lccccccc}
\toprule
\textbf{Method} & \textbf{Vision Encoder} & \textbf{Res.}  & \textbf{GQA} & \textbf{TQA} & \textbf{MMB} & \textbf{MME-P} & \textbf{SEED} \\
\midrule
\multicolumn{8}{l}{\small \textit{w/ Understanding Only}} \\
% Emu3-Chat & MoVQGAN  & 512 & 85.2 & 60.3 & 64.7 & 58.5 & 1243.8 & 68.2 \\
\rowcolor{gray!20} LLaVA-1.5~\cite{llava-1.5} & CLIP ViT-L  & 336  & 62.0 & 46.1 & 64.3 & 1510.7 & 58.6 \\
\rowcolor{gray!20} LLaVA-OV~\cite{llava-ov} & SigLIP ViT-So  & 384 & -- & -- & 80.8 & 1580.0 & 75.4 \\
% InternVL3 & InternViT  & 448 & 91.1 & - & 80.2 & 83.4 & 1748.4 & 77.1 \\
% Qwen2.5-VL~\cite{Qwen-VL} & QwenViT & 448 & 85.9 & - & 84.9 & 83.5 & 1698.1 & 77.0 \\
\midrule
\multicolumn{8}{l}{\small \textit{w/ Unified Tokenizer}} \\
VILA-U~\cite{VILA-U2025} & SigLIP ViT-So  & 384  & 60.8 & 60.8 &- & 1401.8 & 59.0 \\
UniTok~\cite{Unitok} & Vitamin-L  & 256  & - & - & - & 1448.0 & - \\
QLIP~\cite{Qlip} & CLIP ViT-L  & 392  & 61.8 & 55.2 & - & 1498.3 & - \\
% TokenFlow-13B & SigLIP ViT-So & 384 & 86.8 & 62.7 & 61.5 & 68.9 & 1545.9 & 68.7 & 38.7 & 66.7 \\
TokLIP~\cite{TokLIP2025} & SigLIP2 ViT-So  & 384  & 59.5 & - & 67.6 & 1488.4 & 70.4 \\
Tar~\cite{tar} & SigLIP2 ViT-So & 384  & 61.3 & - & 74.4 & \bf 1571.0 & \bf 72.1 \\
VQRAE~\cite{VQRAE2025} & SigLIP2 ViT-So & 512  & \bf 63.6 & 58.8 & 67.6 & 1494.2 & 62.8 \\
\textbf{dRAE} & SigLIP2 ViT-So & 384  & 62.1 & \bf 63.8 & \bf 81.4 & 1468.8 & 71.1 \\
\textbf{dRAE} & SigLIP2 ViT-So & 512  & \bf 62.4 & \textbf{67.7} & \textbf{81.5} & \bf 1535.1 & \textbf{72.7} \\
\bottomrule
\end{tabular}
}
\label{tab:understanding}
\end{table*}

\noindent\textbf{Visual Generation. }
We evaluate the T2I generation performance of dRAE in Tab.~\ref{tab:geneval}. Despite its lightweight architecture ($\sim$700M trainable parameters) and relatively modest training scale, our generative model demonstrates competitive capabilities against baselines trained on significantly larger datasets. This underscores that semantic latents constructed from VFMs serve as a powerful and highly efficient foundation for generative downstream tasks. For C2I generation, models utilizing HSQ-derived tokenizers consistently outperform their VQ-derived counterparts across all codebook sizes, yielding substantial improvements in both gFID and Inception Score (IS), as detailed in Tab.~\ref{tab:c2i_result}. Furthermore, we integrate iREPA~\cite{irepa2025} to enhance internal representation alignment. As shown in Tab.~\ref{tab:repa_result}, the HSQ-based approach benefits more pronouncedly from this alignment strategy. Specifically, while iREPA accelerates early convergence for both tokenizers, the VQ-based baseline suffers from an early performance plateau and noticeable training oscillations.

\begin{table*}[t]\label{tab:gen}
\centering
\small
\caption{\textbf{Evaluation on text-to-image generation}. Data refers to the image-text pairs traversed during training. We leave the exhaustive data and model curation for future study.}
\resizebox{\textwidth}{!}{%
\begin{tabular}{lccccccccc}
\toprule
\multirow{2}{*}{\textbf{Model}}  & \multicolumn{7}{c}{\textbf{GenEval}} & \small \textbf{DPG-Bench} & \multirow{2}{*}{\textbf{Data}}\\
\cmidrule(lr){2-8} \cmidrule(lr){9-9}
   & \textbf{Single} & \textbf{Two} & \textbf{Count} & \textbf{Color} & \textbf{Pos.} & \textbf{Attr.} & \textbf{Overall} & \textbf{Overall} & \\
\midrule
\multicolumn{6}{l}{\textit{Continuous Generation}} \\
SDv1.5~\cite{SD-VAE2022}  & 0.97 & 0.38 & 0.35 & 0.76 & 0.04 & 0.06 & 0.43 & 63.18 & $>1$B\\
DALL-E 3~\cite{dalle3}  & 0.96 & 0.87 & 0.47 & 0.83 & 0.43 & 0.45 & 0.67 & 83.50 & $>1$B\\
% Scale-RAE~\cite{scale-rae}  & - & - & - & - & - & - & 0.50 & 76.90 & 90M \\
% SD3.5-M~\citet{esserScalingRectifiedFlow2024}  & 0.99 & 0.94 & 0.72 & 0.89 & 0.33 & 0.60 & 0.74 & 84.08 & xx \\
\midrule
\multicolumn{6}{l}{\textit{Discrete Generation}} \\
LlamaGen~\cite{llamagen}  & 0.71 & 0.34 & 0.21 & 0.58 & 0.07 & 0.04 & 0.32 & 64.84 & 60M \\
TokenFlow~\cite{tokenflow}  & -- & -- & -- & -- & -- & -- & 0.55 & 73.38 & 60M \\
Show-o~\cite{show-o}  & 0.95 & 0.52 & 0.49 & 0.82 & 0.11 & 0.28 & 0.53 & -- & 35M \\
Meissonic~\cite{meissonic}  & 0.99 & 0.66 & 0.42 & 0.86 & 0.10 & 0.22 & 0.54 & -- & $>$200M \\
Janus~\cite{janus}  & 0.97 & 0.68 & 0.30 & 0.84 & 0.46 & 0.42 & 0.61 & 79.68 & $>$100M \\
% MMaDA~\citet{yangMMaDAMultimodalLarge2025}  & 0.96 & 0.60 & 0.45 & 0.81 & 0.14 & 0.25 & 0.56 & 70.51\\
\textbf{dRAE} & 0.98 & 0.69 & 0.45 & 0.92 & 0.30 & 0.39 & \bf 0.63 & \bf 80.58 & \bf 12M \\
\bottomrule
\end{tabular}%
}
\label{tab:geneval}
\end{table*}

\section{Ablative Study}\label{sec:ablation}

\begin{table*}[t]
\centering
\begin{minipage}{0.5\linewidth}
\centering
% ================= Table 5 =================
\caption{Ablation on C2I generation. Using simplest sampler without hyperparameter searching.}
\label{tab:c2i_result}
\begin{tabular}{lcccc}
\toprule
 Method & \small Codebook & \small rFID & \small gFID$\downarrow$ &  \small IS$\uparrow$ \\
\midrule
 VQ & 32768 & 2.25 & 5.37 & 251.3 \\
 VQ & 65536 & 2.21 & 5.51 & 264.3 \\
 HSQ & 32768& 2.20 & 4.83 & 268.4  \\
 HSQ & 65536 & 2.14 & 4.45 & 287.3  \\
\bottomrule
\end{tabular}
\end{minipage}
\hfill
% ================= Table 6 =================
\begin{minipage}{0.45\linewidth}
\centering
\caption{Ablation on alignment. iREPA for improved representation alignment~\cite{irepa2025}.}
\label{tab:repa_result}
\begin{tabular}{lccc}
\toprule
Method & Configuration & \small gFID$\downarrow$ & \small IS$\uparrow$ \\
\midrule
VQ & w/o iREPA & 6.89 & 215.3 \\
VQ & w/ iREPA   & 7.16 & 226.1\\
HSQ & w/o iREPA & 6.65 & 251.4\\
HSQ & w/ iREPA &  6.11 & 256.3\\
\bottomrule
\end{tabular}
\end{minipage}
\vfill
\vspace{0.3cm}
% ================= Table 7 =================
\begin{minipage}{0.5\linewidth}
\centering
\caption{Ablation of objective metric on reconstruction tasks. $\ell_2$ for standard Eucildean metric, and $\theta$ for cosine-similarity based method.}
\label{tab:ablation_rec}
\begin{tabular}{ccccc}
\toprule
 Assign &  \small $\mathcal{L}_{\text{codebook}}$ & \small $\mathcal{L}_{\text{commit}}$ &  \small rFID  & \small SSIM \\
\midrule
 $\ell_2$ & $\ell_2$ & $\ell_2$ & 3.59 & 0.62 \\
 $\theta$ & $\ell_2$ & $\ell_2$ & 3.31 & 0.62 \\
 $\theta$ & $\theta$ & $\ell_2$ & \textbf{3.02} & \textbf{0.65} \\
 $\theta$ & $\theta$ & $\theta$ & 12.7 & 0.33 \\
\bottomrule
\end{tabular}
\end{minipage}
\hfill
% ================= Table 8 =================
\begin{minipage}{0.45\linewidth}
\centering
\caption{Ablation of tokenizer design on multimodal understanding tasks under identical visual instruction tuning setting.}
\label{tab:ablation_und}
\begin{tabular}{lccc}
\toprule
Method & Codebook & GQA & MMB \\
\midrule
VQ & 16384   & 34.1 & 43.8 \\
VQ & 65536   & 33.8 & 42.0 \\
HSQ & 16384  & 34.3 & 44.1 \\
HSQ & 65536  & 35.8 &  45.6 \\
\bottomrule
\end{tabular}
\end{minipage}

\end{table*}

% \begin{table*}[t]
% \centering
% % ================= Table 5 =================
% \begin{minipage}{0.5\linewidth}
% \centering
% \caption{Results for Feature Reconstruction}
% \label{tab:add_feature_rec}
% \begin{tabular}{ccccc}
% \toprule
%  Method & \small Codebook & \small Util. &  \small PSNR  & \small Cos. Sim. \\
% \midrule
%  VQ & 32768 & -- & -- & -- \\
%  VQ & 65536 & -- & -- & -- \\
%  HSQ & 32768& -- & \textbf{--} & \textbf{--} \\
%  HSQ & 65536 & -- & \textbf{--} & \textbf{--} \\
% \bottomrule
% \end{tabular}
% \end{minipage}
% \hfill
% % ================= Table 6 =================
% \begin{minipage}{0.45\linewidth}
% \centering
% \caption{Results for Image Generation.}
% \label{tab:add_und_rec}
% \begin{tabular}{lccc}
% \toprule
% Method & Codebook & GQA & MMB \\
% \midrule
% VQ & 16384   & 34.1 & 43.8 \\
% VQ & 65536   & 33.8 & 42.0 \\
% \bf HSQ & 16384  & 34.3 & 44.1 \\
% \bf HSQ & 65536  & 35.8 &  45.6 \\
% \bottomrule
% \end{tabular}
% \end{minipage}

% \end{table*}

\noindent\textbf{Loss Design.}
We evaluate the contribution of each component in our design by systematically ablating on the routing mechanism and loss formulations. Table~\ref{tab:ablation_rec} summarizes the evaluated variants. The baseline (line 1) uses Euclidean routing with $\ell_2$-based codebook and commitment loss. Replacing Euclidean distance with cosine similarity (line 2) already improves performance, indicating the benefit of directional matching in the latent space. Building on this, our method (line 3) introduces a cosine-based $\mathcal{L}_{\text{codebook}}$ while retaining an $\ell_2$ commitment loss. The full-spherical variant (line 4) applies cosine metrics to all components. 
% All methods use a projection layer upon the codebook embeddings as SimVQ for efficient training and fair comparison.

The results show that the best performance is achieved by combining a cosine-based $\mathcal{L}_{\text{codebook}}$ with an $\ell_2$ commitment loss. 
This allows the codebook to align with the directional structure of the latent space, while the Euclidean commitment term preserves feature magnitude and guides the projection to meet with codebook's distribution. 
When both losses are defined in cosine space, performance degrades significantly, suggesting that magnitude information remains important for convergence. We also provide additional experiment results with DINOv2 as the encoder in \ref{ablate:dino}.

\noindent\textbf{Impact on Semantics.}
We further explore the use of quantized features for training multimodal models and evaluate the impact of different codebook designs on downstream understanding performance. As shown in Tab~\ref{tab:ablation_und}, performance with HSQ consistently improves as the codebook size increases, whereas VQ-based methods exhibit relatively stagnant behavior. This observation further supports our claim that semantically aligned encoding leads to superior performance.In practice, we also observe that VQ-based approaches suffer from notable instability during training: the reconstruction loss exhibits persistent oscillations, while the distillation loss tends to undergo an irreversible increase in the later stages of training, indicating a growing deviation from the pretrained teacher model. Such issues are not observed with HSQ. 
% We hope these findings provide useful insights for the development of more unified and fully end-to-end trainable tokenizers in the future.
% \paragraph{Effect of Feature Normalization.}
% The \emph{Pre-norm VQ} variant performs significantly worse, indicating that explicitly normalizing features before quantization is detrimental. While angular routing benefits from scale invariance, removing magnitude information from the latent representation reduces the decoder's capacity to model fine-grained details, leading to degraded reconstruction quality.

\section{Beyond Image Reconstruction}

The main experiments in this paper follow the RAE paradigm, where image reconstruction serves as a \textit{proxy task} for training the tokenizer. Nevertheless, the core idea of HSQ is to learn an effective mechanism for quantizing high-dimensional representations, which should be applicable regardless of the choice of proxy task. 
% For instance, a natural question is whether the decoder can directly reconstruct quantized high-dimensional features. We are also interested in how HSQ performs when the codebook is trained end-to-end together with modern VLMs. 
Several recent works~\cite{gengXOmniReinforcementLearning2025, dingKelixTechnicalReport2026,teamLongCatNextLexicalizingModalities2026,wangRepresentationForcingBottleneckFree2026,pengUnifiedMultimodalAutoregressive2026} have discussed related directions. We explore alternative proxy and report two preliminary experiments:

\begin{minipage}{0.66\linewidth}

\noindent\textbf{Feature Reconstruction.}
With the encoder $E$ frozen, we get the encoded image features $z=E(x)$ and train a codebook together with a ViT decoder to reconstruct the features $\hat z$. The reconstruction objective is defined as $\mathcal{L}_{\text{rec}}=\text{L}_2(\hat{z},z)$. We train the quantizer and decoder for 100K steps and evaluate the PSNR and cosine similarity between $z$ and $\hat z$. As shown in Tab.~\ref{tab:feat_recon}, HSQ significantly outperforms VQ. Nevertheless, reconstructing high-dimensional patch embeddings is more challenging than reconstructing pixel values, which we attribute to the intrinsic noise in the encoder representations and the absence of perceptual supervision~\cite{zhangUnreasonableEffectivenessDeep2018}, from which VAEs benefit.
\end{minipage}%
\hfill
\begin{minipage}{0.32\linewidth}
    \centering
    \captionof{table}{Comparison of VQ and HSQ on feature reconstruction.}
    \label{tab:feat_recon}
    \small
    \begin{tabular}{lcc}
        \toprule
        Method & PSNR $\uparrow$ & Sim. $\uparrow$ \\
        \midrule
        VQ & 4.91 & 0.81 \\
        \bf HSQ & \bf 8.24 & \bf 0.92 \\
        \bottomrule
    \end{tabular}
\end{minipage}

\noindent\textbf{Joint Image-Semantic Modeling.}
We insert the quantizer between the frozen visual encoder $E$ and the trainable language backbone $D_{\text{LLM}}$ of Qwen-3.5~\cite{qwen3.5}, while attaching a trainable ViT decoder $D_{\text{pixel}}$ in parallel. The codebook is then trained end-to-end using both image reconstruction and image understanding objectives, allowing supervision to flow through a unified training objective. Let $z_q$ denote the quantized image features, the reconstructed image $\hat x = D_{\text{pixel}}(z)$, and $y$ as the language response for query $q$ about image $x$. We use 4M image-query-response pairs $\{x,q,y\}$ from LLaVA-OneVision~\cite{llava-ov} as the training dataset. The overall objective combines the language modeling loss and the image reconstruction loss as $\mathcal{L}_{\text{rec}}=-\sum_t \log p_{D_{\text{LLM}}}\!\left(y_t \mid y_{<t},q, z_q\right) + \text{L}1_{D_{\text{pixel}}}(\hat{x}, x)$.

Through experiments, we found that an entropy regularization loss~\cite{changMuseTextToImageGeneration2023,yuLanguageModelBeats2024a} is essential for effective codebook utilization when an LLM is used as the decoder. Since this loss is already included in the IBQ family, we plug it to HSQ and VQ for a fair comparison. We compare HSQ with VQ, IBQ~\cite{shiScalableImageTokenization2025}, and the $\ell_2$-normalized variant of IBQ~\cite{gengXOmniReinforcementLearning2025}, with a vocabulary size of $K=131{,}072$. The entropy regularization uses a temperature-scaled softmax with temperature $\tau=0.01$ for all methods.

\begin{figure}
    \centering
    \includegraphics[width=1.0\linewidth]{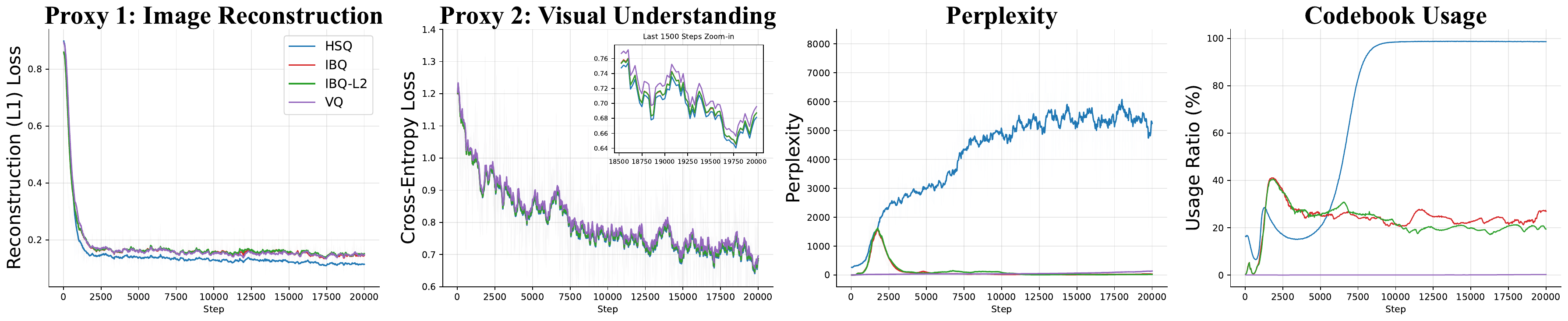}
    \caption{Comparison of quantization methods under the joint image-semantic optimization.}
    \label{fig:semantic-ntp}
\end{figure}

As shown in Fig.~\ref{fig:semantic-ntp}, VQ performs the worst among all compared methods, exhibiting slow convergence and poor codebook usage. The IBQ variants benefit from stochastic sampling and consequently achieve higher code utilization. HSQ, despite not relying on randomized sampling, achieves the highest codebook utilization and the fastest convergence in both image reconstruction and understanding tasks by simply adopting angular-based routing and codebook updates. These results further demonstrate the effectiveness of HSQ for quantizing high-dimensional representations.

\section{Conclusion}
We presented dRAE, a Representation Autoencoder with Hyper-Spherical Codes, directly tackling the pervasive codebook collapse issue in high-dimensional discrete visual tokenizers. By aligning the codebook update dynamics more closely with the underlying semantic latent distribution, our approach achieves exceptionally high codebook utilization, fast convergence and robust reconstruction quality with minimal anti-collasping tricks. This study provides a fresh insight for training visual tokenizers, as well as laying a scalable foundation for training next-generation MLLMs.

\section{Acknowledgments}
% This work was supported by National Natural Science Foundation of China under Grant 62225208, 62521007 and CAS Project for Young Scientists in Basic Research under Grant No.YSBR-117.

This work was supported by Ant Group Research Intern Program.

{
\small
\bibliographystyle{plain}
\bibliography{ref}
}

\appendix
\section{Implementation Details}\label{appen:details}

\subsection{Datasets} 
dRAE is trained to reconstruct the the open-sourced 36M images of BLIP3-o~\cite{blip3o}, following UniLIP~\cite{unilip} and VQRAE~\cite{VQRAE2025}. 
% The ablation studies, unless specified, are also conducted on the same data. 
For image understanding, we follow the alignment and pretraining setting of LLaVA-1.5~\cite{llava-1.5}, and conduct instruction tuning~\cite{llava-ov}.
% which uses LLaVA-Pretrain-595K and LLaVA-v1.5-mix665K in ablation, and utilize . 
For T2I generation, we use 12M image-text pairs, including Megalith-10M~\cite{BoerBohan2024Megalith10m} and T2I-2M~\cite{zou2026advancing} for generative pretraining, and BLIP3-o-60K~\cite{blip3o} for instruction fine-tuning.

\subsection{Tokenizer Training}

\begin{figure}
    \centering
    \includegraphics[width=1.0\linewidth]{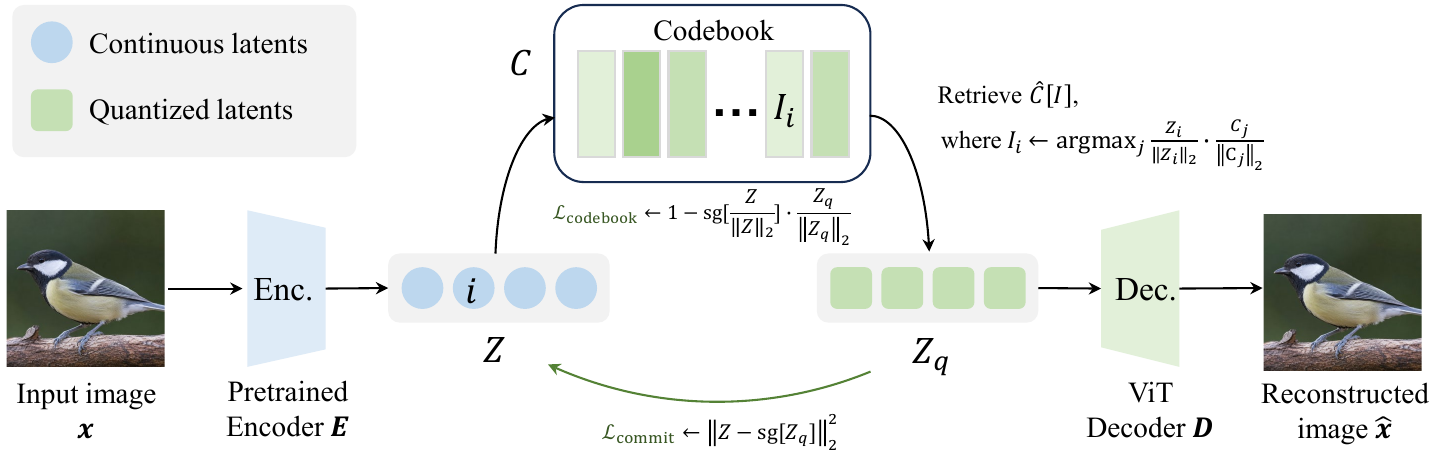}
    \caption{A demonstration of optimizing the codebook embeddings, following VQ-VAE~\cite{VQVAE2017}.}
    \label{fig:pipe}
\end{figure}

\paragraph{Algorithm}
The complete training algorithm for HSQ is demonstrated in Alg.~\ref{alg:spherical_vq}. We also provide an illustration in Fig~\ref{fig:pipe}. The training is conducted on $8\times 80\text{G}$ NVIDIA GPUs for 3 days.

\begin{algorithm}[h]
\caption{Hyper-Spherical Quantization}
\label{alg:spherical_vq}
\begin{algorithmic}[1]

\STATE{\textbf{Input: }Feature tensor $Z \in \mathbb{R}^{N\times d}$. 
Base codebook embeddings $\mathcal{C}$, learnable projection $W$, and commitment weight $\beta$} \\
\vspace{0.3em}
\STATE $\hat{C} \gets W\mathcal{C}$
\algcomment{Project codebook following SimVQ}\\
\STATE $S_{i,j} \gets \frac{Z_i}{\|Z_i\|_2 } \cdot \frac{\hat C_j}{ \|\hat C_j\|_2}$
\algcomment{Compute cosine similarity}\\
\STATE $I_i \gets \arg\max_j S_{i,j}$
\algcomment{Angular routing}\\
\STATE $Z_q \gets \hat{C}[I]$
\algcomment{Lookup on projected codebook}\\
\STATE $\mathcal{L}_{\text{codebook}} \gets 1-\text{sg}[\frac{Z}{\|Z\|_2 }] \cdot \frac{Z_q}{ \|Z_q\|_2}$
\algcomment{Angular codebook loss}\\
\STATE $\mathcal{L}_{\text{commit}} \gets \|Z - \text{sg}[Z_q]\|^2_2$
\algcomment{Commitment loss}\\
\STATE $\mathcal{L}_{\text{VQ}} \gets \mathcal{L}_{\text{codebook}} + \beta \mathcal{L}_{\text{commit}}$
\STATE $Z_{\text{out}} \gets Z + \text{sg}[Z_q - Z]$
\algcomment{Straight-through estimator}\\
\vspace{0.5em}

\STATE{\textbf{Return: }$Z_{\text{out}}, \mathcal{L}_{\text{VQ}}, I$}
\end{algorithmic}
\end{algorithm}

\paragraph{Hyperparameter}
The training details of dRAE are provided in Tab.~\ref{tab:training_hyperparams}. 
% Due to limited time and computational budget, we did not perform extensive hyperparameter optimization. Additional gains might be obtained through more careful tuning. 
The perceptual loss design follows the implementation of VQRAE, while the discriminator design follows RAE.

\begin{table*}[h]
\centering

\begin{minipage}{0.5\linewidth}
\centering
\caption{Ablation on reconstruction tasks with DINOv2-B Encoder.}
\label{tab:ablation_rec_dinov2_1}
\begin{tabular}{ccccc}
\toprule
 Assign &  \small $\mathcal{L}_{\text{codebook}}$ & \small $\mathcal{L}_{\text{commit}}$ &  \small rFID  & \small SSIM \\
\midrule
 $L_2$ & $L_2$ & $L_2$ & 3.67 & 0.56 \\
 $\theta$ & $L_2$ & $L_2$ & 4.23 & 0.57 \\
 $\theta$ & $\theta$ & $L_2$ & \textbf{2.87} & \textbf{0.58} \\
\bottomrule
\end{tabular}
\end{minipage}
\hfill
% ================= Table 6 =================
\begin{minipage}{0.45\linewidth}
\centering
\caption{Ablation on reconstruction tasks with RAEv2-style DINOv2-B Encoder.}
\label{tab:ablation_rec_dinov2_2}
\begin{tabular}{ccccc}
\toprule
 Assign &  \small $\mathcal{L}_{\text{codebook}}$ & \small $\mathcal{L}_{\text{commit}}$ &  \small rFID  & \small SSIM \\
\midrule
 $L_2$ & $L_2$ & $L_2$ & 4.24 & 0.56 \\
 $\theta$ & $L_2$ & $L_2$ & 4.11 & 0.58 \\
 $\theta$ & $\theta$ & $L_2$ & \textbf{3.29} & \textbf{0.60} \\
\bottomrule
\end{tabular}
\end{minipage}

\vspace{0.5em}

% ================= Table 10 =================
\begin{minipage}{0.48\linewidth}
\centering
\small
\caption{Details for tokenizer training.}
\label{tab:training_hyperparams}
\resizebox{\textwidth}{!}{%
\begin{tabular}{llc}
\toprule
\textbf{Category} & \textbf{Hyperparameter} & \textbf{Value} \\
\midrule
\multirow{3}{*}{Architecture} & Image Encoder & SigLIP2-So400M \\
 & Encoder Input Size & 512 \\
 & Decoder Type & ViT-XL \\
\midrule
\multirow{7}{*}{Training} & Global Steps & 150000 \\
 & Global Batchsize & 512 \\
 & Optimizer & AdamW ($0.9, 0.95$) \\
 & Learning Rate  & $2.0 \times 10^{-4}$ \\
 & Encoder Learning Rate & $2.0 \times 10^{-5}$ \\
 & Weight Decay & 0.0 \\
 & EMA Decay & 0.9978 \\
\midrule
\multirow{3}{*}{LR Scheduler} & Type & Cosine \\
 & Warmup Epochs & 1 \\
 & Final Learning Rate & $2.0 \times 10^{-5}$ \\
\midrule
\multirow{3}{*}{Discriminator} & Disc. Learning Rate & $5.0 \times 10^{-5}$ \\
 & Disc. Loss Type & Hinge \\
 & Gen. Loss Type & Vanilla \\
\midrule
\multirow{3}{*}{Loss Weights} & Perceptual Weight $\omega_{\text{p}}$ & 1.0 \\
 & Discriminator Weight $\omega_{\text{d}}$ & 0.1 \\
 & Distillation Weight $\lambda$ & 1.0 \\
\bottomrule
\end{tabular}
}
\end{minipage}
\hfill
% ================= Table 6 =================
\begin{minipage}{0.48\linewidth}
\centering
\small
\caption{Details for generative models.}
\label{tab:gen_training_hyperparams}
\resizebox{\textwidth}{!}{%
\begin{tabular}{llc}
\toprule
\textbf{Category} & \textbf{Hyperparameter} & \textbf{Value} \\
\midrule
\multirow{4}{*}{Architecture} & Text Encoder & T5-XXL \\
 & Max Text Length & 64 \\
 & Image Resolution & 384 \\
 & Decoder Type & DiT-H \\
 & Codebook & 65536 \\
\midrule
\multirow{7}{*}{Training} & Global Steps & 125000 \\
 & Global Batchsize & 512 \\
 & $t$ Distribution & Logit Norm \\
 & Optimizer & AdamW ($0.9, 0.95$) \\
 & Learning Rate  & $2.0 \times 10^{-4}$ \\
 & Weight Decay & 0.0 \\
 & EMA Decay & 0.9999 \\
\midrule
\multirow{2}{*}{LR Scheduler} & Type & Constant \\
 & Warmup Epochs & 1 \\
\midrule
\multirow{3}{*}{Inference} & Steps & $50$ \\
 & Schedule & Cosine \\
 & CFG & $3.5$ \\
\bottomrule
\end{tabular}
}
\end{minipage}

\end{table*}

\subsection{Additional Results}\label{ablate:dino}

We set a variant of dRAE at $256\times256$ resolution, using DINOv2-B~\cite{DINOv2} as the encoder, a 12 layer ViT as the decoder, and train it for $20000$ steps with 512 global batch size. The ablation results are shown in Tab.~\ref{tab:ablation_rec_dinov2_1}. We also follow RAEv2~\cite{singh2026raev2} and test a variant that adds multiple layer features of the encoder for better reconstruction quality. Specifically, we use the sum of hidden states from DINOv2's \texttt{\{3,6,9,12\}} layers. The corresponding results are shown in Tab.~\ref{tab:ablation_rec_dinov2_2}. These experiments indicate that HSQ can also effectively capture the distribution of self-supervised visual encoders.

\subsection{MLLM Training}\label{appn:mllm}

\paragraph{Data usage and fair comparison}
For the main results table, the MLLM is trained using a combination of the publicly available datasets from LLaVA-1.5 and 1M randomly sampled image-text pairs from LLaVA-OneVision. We additionally train a controlled variant using a 384-resolution encoder, Vicuna-7B as the LLM backbone, and only the LLaVA-1.5 data. The comparison with other academic works is shown in Tab.~\ref{appen-tab:und}. Under this more standardized training setup, our model—built upon an encoder fine-tuned for reconstruction—still demonstrates advantages across multiple comprehension metrics, further validating the potential of dRAE as a unified tokenizer.

\begin{table*}[t]
\centering
\small
\caption{\textbf{Evaluation on multimodal understanding}. All models use Vicuna-7B as the backbone and only LLaVA-1.5 data.}
\label{appen-tab:und}
\resizebox{\textwidth}{!}{%
\begin{tabular}{lccccccc}
\toprule
\textbf{Method} & \textbf{Vision Encoder} & \textbf{Res.} & \textbf{POPE} & \textbf{GQA} & \textbf{TQA} & \textbf{MMB} & \textbf{SEED} \\
\midrule
LLaVA-1.5~\cite{llava-1.5} & CLIP ViT-L  & 336 & 85.9 & 62.0 & 46.1 & 64.3  & 58.6 \\
% InternVL3 & InternViT  & 448 & 91.1 & - & 80.2 & 83.4 & 1748.4 & 77.1 \\
VILA-U~\cite{VILA-U2025} & SigLIP ViT-So  & 384 & 85.8 & 60.8 & 60.8 &-  & 59.0 \\
UniTok~\cite{Unitok} & Vitamin-L  & 256 & 81.7 & - & - & -  & - \\
QLIP~\cite{Qlip} & CLIP ViT-L  & 392 & 86.1 & 61.8 & 55.2 & -  & - \\
TokLIP~\cite{TokLIP2025} & SigLIP2 ViT-So  & 384 & 84.1 & 59.5 & - & 67.6  & 70.4 \\
\textbf{dRAE} & SigLIP2 ViT-So & 384 & 85.5 & 61.0 & 60.1 & 71.7  & 65.0 \\
\bottomrule
\end{tabular}
}
\end{table*}

\subsection{Generative Model Training}\label{appen:training}

\paragraph{Discrete Diffusion Model (DDM)}

DDM defines a forward process on discrete variables by gradually corrupting tokens to absorbing state $\mathbf{m}$ through a continuous-time Markov process~\cite{sahooSimpleEffectiveMasked2024}.
We denote clean data as $x_{t=0}$ ($x_0$ for short), and noise it gradually as $t\rightarrow 1$.
Let \( \alpha_t \) be the noise scheduler (a monotonically decreasing survival function that satisfies \(\alpha_0=1, \alpha_1=0\) ), the corrupted data distribution at time $t$ is determined as 
\begin{equation}\label{denote}
    x_t \sim q(x_t|x_0, t), q(x_t|x_0, t) = \text{Cat}(x_t; \alpha_t x_0 + (1-\alpha_t)\mathbf{m}).
\end{equation}
Let $\delta(x_{(t,i)},\textbf{m})$ be the indicator function that is activated only if the $i$-th position of $x_t$ is $\mathbf{m}$.
For a linear scheduler, the objective is derived as the evidence lower bound (ELBO) of $\log \pi_\theta(x_0|x_t)$:
\begin{equation}\label{ddmloss}
    \mathcal{L}_{\text{DDM}} = -\mathbb{E}_{t,\ x_0,\ x_t} [\frac{1}{t}\sum_{i=1}^L\delta(x_{(t,i)},\mathbf{m})\log \pi_\theta(x_{(0,i)}|x_t)]=-\mathbb{E}_{t,\ x_0,\ x_t} [\ell_{\pi_\theta}(x_t,x_0)].
\end{equation}
For conditional generation where a prompt $\mathbf{c}$ is given, we write $\ell_{\pi_\theta}(x_t,x_0|\mathbf{c})$ for simplicity.
Following MDLM's deduction, assume that the network can reconstruct $x_0$ perfectly, we use $\pi_\theta(x_t)$ to approximate this denoising process, and get the sampling rule as
\begin{equation}\label{reverse}
    p_\theta(x_s|x_t)= 
    \begin{cases}
    1, & \text{if } x_s = x_t, \ x_t \ne \mathbf{m} ,\\
    \frac{1-\alpha_s}{1-\alpha_t}, & \text{if } x_s =\mathbf{m}, \ x_t =\mathbf{m},\\
    \frac{\alpha_s-\alpha_t}{1-\alpha_t}\pi_\theta(x_t), & \text{if } x_s \ne \mathbf{m}, \ x_t = \mathbf{m},  \\
    0, & \text{otherwise.} \\
    \end{cases}
\end{equation}

\paragraph{Hyperparameter}

The details of T2I generative training are provided in Tab.~\ref{tab:gen_training_hyperparams}. Since a resolution of 512 requires 1024 tokens per image, the batch size becomes constrained on each machine. Therefore, we additionally train a dRAE with SigLIP2 ViT-large-patch16-384, which compresses each image into 576 tokens, enabling more efficient generative training. In addition, since a large vocabulary also significantly increases the memory footprint of the output layer, we set the tokenizer vocabulary size to 65,536. The training is conducted on 16 GPUs for 3 days.

For C2I generation, we use SigLIP2 ViT-base-patch16-256 as the visual encoder, and train the tokenizer at 256 resolution, which means compressing each image into 256 tokens. The generative model is a 12-layer transformer. The results in Tab.~\ref{tab:c2i_result} are gained after training 300 epochs on ImageNet. The ablation in Tab.~\ref{tab:repa_result} are gained with 65536-vocab tokenizers after 100 epochs, using the default iREPA design for B-size DiT models.

\subsection{Evaluation}\label{appen:eval}

\paragraph{Reconstruction}
We follow the evaluation protocol of RAE and assess reconstruction performance on the ImageNet validation set. During the reproduction of prior work, we observed that the SSIM evaluation in UniLIP contains a bug: the \texttt{data-range} passed to the evaluator is incorrectly set to $2.0$ instead of the correct value of $1.0$ corresponding to the actual image dynamic range, leading to an overestimation of SSIM by $\sim 10\%$. The VQRAE implementation, based on the same codebase, suffers from the same issue. Accordingly, we correct the reported SSIM values for both methods based on our experiments.

\paragraph{Understanding}
We use the official script in the repository of LLaVA-1.5 for evaluation.

\paragraph{Generation}
We follow the standard protocol of GenEval and DPG-Bench, which samples 4 images for each prompt and use their evaluation model to score. The images are sampled with 50 steps using a cosine schedule and a constant classifier free guidance (CFG) of 3.5.

\end{document}